\documentclass[letterpaper]{article} 
\usepackage{aaai25}  
\usepackage{times}  
\usepackage{helvet}  
\usepackage{courier}  
\usepackage[hyphens]{url}  
\usepackage{graphicx} 
\usepackage{natbib}  
\usepackage{caption} 
\usepackage{algorithm}
\usepackage{algorithmic}

\usepackage{amsmath,amsfonts}
\usepackage{array}
\usepackage[caption=false,font=normalsize,labelfont=sf,textfont=sf]{subfig}
\usepackage{textcomp}
\usepackage{url}
\usepackage{verbatim}
\usepackage{multirow}
\usepackage{cite}
\usepackage{amssymb}
\usepackage{bbding}
\usepackage{url}

\usepackage{newfloat}
\usepackage{listings}
\DeclareCaptionStyle{ruled}{labelfont=normalfont,labelsep=colon,strut=off} 
\floatstyle{ruled}
\newfloat{listing}{tb}{lst}{}
\floatname{listing}{Listing}
\title{4DStyleGaussian: Zero-shot 4D Style Transfer with Gaussian Splatting}
\author{
    Wanlin Liang\equalcontrib,
    Hongbin Xu\equalcontrib,
    Weitao Chen,
    Feng Xiao,
    Wenxiong Kang\textsuperscript{\rm}\thanks{Corresponding author.}
}
\affiliations{

}

\usepackage{bibentry}

\begin{document}

\maketitle

\begin{abstract}

3D neural style transfer has gained significant attention for its potential to provide user-friendly stylization with spatial consistency. However, existing 3D style transfer methods often fall short in terms of inference efficiency, generalization ability, and struggle to handle dynamic scenes with temporal consistency. In this paper, we introduce 4DStyleGaussian, a novel 4D style transfer framework designed to achieve real-time stylization of arbitrary style references while maintaining reasonable content affinity, multi-view consistency, and temporal coherence. Our approach leverages an embedded 4D Gaussian Splatting technique, which is trained using a reversible neural network for reducing content loss in the feature distillation process. Utilizing the 4D embedded Gaussians, we predict a 4D style transformation matrix that facilitates spatially and temporally consistent style transfer with Gaussian Splatting. Experiments demonstrate that our method can achieve high-quality and zero-shot stylization for 4D scenarios with enhanced efficiency and spatial-temporal consistency. 
\end{abstract}

\section{Introduction}

Neural artistic stylization applies neural networks to create a new image that retains the structural elements of the content image but adopts the stylistic features of the style image, which has seen significant advancements recently\cite{gatys2016image,johnson2016perceptual,huang2017arbitrary,li2017universal,sheng2018avatar,li2019learning,park2019arbitrary,deng2020arbitrary,liu2021adaattn}. Benefiting from the emerging implicit neural representation, such as Neural Radiance Fields(NeRFs)\cite{li2022neural}, 3D style transfer methods\cite{huang2021learning,huang2022stylizednerf,chiang2022stylizing,nguyen2022snerf,zhang2022arf,fan2022unified,liu2023stylerf,jung2024geometry,chen2024upst} achieve outstanding performance with multi-view consistency. Nonetheless, NeRF-based stylization methods have limitations in memory and time consumption, for they are computationally intensive to optimize and infer, which hinders instant stylizations. Additionally, the original radiance fields may cause visually unpleasant geometry artifacts in the stylization results. It is crucial that the 3D style transfer methods can be capable of rendering stylized novel views in real time without sacrificing good visual effects.

Recently, 3D Gaussian Splatting(3D-GS)\cite{kerbl20233d} has gained increasing attention for its rapid rendering capabilities, making real-time 3D style transfer possible. Existing 3DGS-based style transfer methods have made remarkable advancements in synthesizing high-quality stylized results in real-time by optimizing a stylized Gaussian Splatting\cite{saroha2024gaussian,zhang2024stylizedgs} or by training embedded Gaussains which are used for style transfer in the feature space\cite{liu2024stylegaussian}. However, these methods still have some limitations: 1) Existing photorealistic methods typically employ a pre-trained encoder and decoder(e.g. VGG\cite{simonyan2014very} which is specially designed for object-level classification, inevitably causing content loss and artifacts in the stylization results. 2) Most of the methods are focused on 3D static scenes, failing to stylize the 4D dynamic scenes, and making it hard to maintain multi-view and cross-time consistency. 

\begin{figure}[!t]
\centering
\includegraphics[width=0.8\columnwidth]{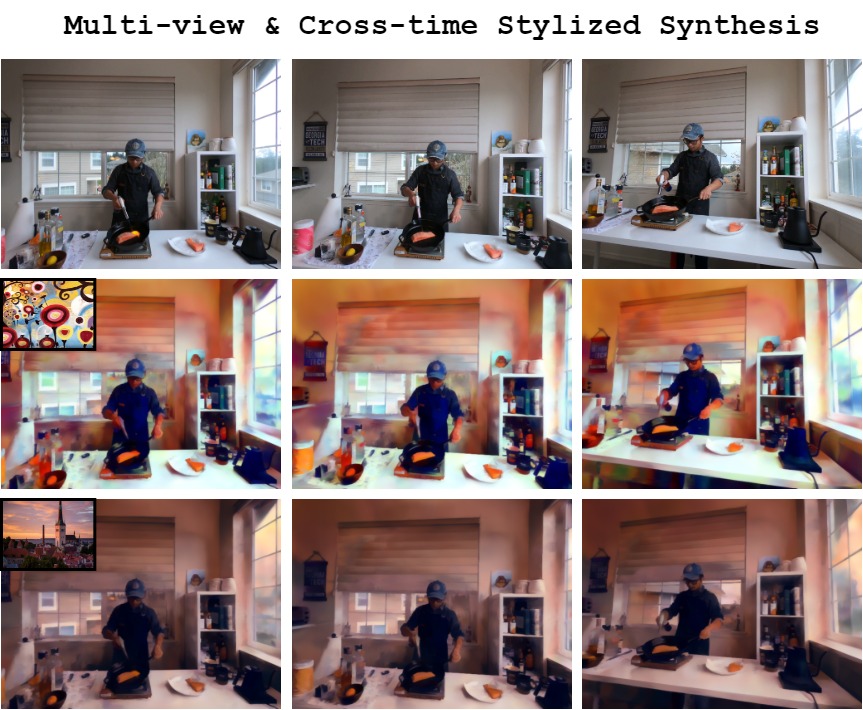}
\vspace{-0.3cm}
\caption{\textbf{Zero-shot 4D Style Gaussian.} 4DStyleGaussian can transfer the reference style to the 4D scene in a zero-shot manner and maintain multi-view and cross-time consistency.}
\vspace{-0.6cm}
\label{introduction}
\end{figure}

In this paper, we tackle these challenges and introduce 4DStyleGaussian, a novel 4D style transfer method that utilizes Gaussian Splatting to achieve instant, high-quality zero-shot stylization with strong content affinity and robust spatial-temporal consistency(As shown in Fig.\ref{introduction}). For the first challenge, we optimize an embedded Gaussian Splatting with a learnable reversible neural network. Specifically, based on the 4D-GS\cite{wu20244d} backbone, we propose to learn an embedding feature at each Gaussian, and by differentiable splatting and rasterizing, we can get the embedded feature maps which are then fed into two MLPs to obtain color maps and feature maps respectively. Innovatively, we apply a 2D reversible neural network that can extract features from content images without content affinity loss to supervise the rendered feature map. Taking advantage of the bijective transformation from the reversible neural network, we can significantly decrease the loss of content information and artifacts caused by the traditional encoder-decoder framework. For the second challenge, we aim to optimize a 4D style transformation matrix to ensure spatial-temporal consistency in the stylization of 4D scenarios. With the pre-trained embedded Gaussians and style images, we predict the whitening and coloring transformation matrix in \cite{li2017universal} through a lightweight MLP and CNN. This 4D linear style transformation matrix notably perceives the spatial and temporal information from 4D Gaussians and reflects a straightforward matching of feature covariance of the embedded Gaussians to a given style image.

In summary, the contributions of our method are as follows:
\begin{itemize}
\item We propose 4DStyleGaussian, the first zero-shot 4D style transfer method with Gaussian Splatting, which can efficiently synthesize stylized novel views of every style reference and maintain multi-view and cross-time consistency.

\item We design a 4D embedded Gaussian Splatting based on a reversible neural network for preserving content affinity and reducing the artifacts in the stylization.

\item We optimize a 4D style transformation matrix based on the pre-trained embedded Guassians and style images to ensure spatial and temporal consistency.  

\item Extensive experiments show that our method achieves promising stylization results in 4D dynamic scenarios.
\end{itemize}

\section{Related Work}
\subsection{Image and Video Stylization} Style transfer methods tend to endow the content image with artistic patterns from the given style image while maintaining the original content information. \cite{gatys2016image} firstly utilizes a Convolutional Neural Network(CNN) to separate and recombine the image content and style of natural images. However, it requires an iterative training process that limits its performance. For faster training, subsequent works\cite{huang2017arbitrary,li2017universal,sheng2018avatar,park2019arbitrary,liu2021adaattn} apply feed-forward neural networks, such as AdaIN\cite{huang2017arbitrary}, which introduces a novel adaptive instance normalization(AdaIN) layer that aligns the mean and variance of the content features with the style features. Video stylization is another topic that focuses on obtaining stylization consistency between adjacent frames. Most Methods\cite{chen2017coherent,huang2017real,ruder2018artistic,wang2020consistent} are based on optical flow or other temporal constraint to existing style transfer methods. CAP-VSTNet\cite{wen2023cap} utilizes Cholesky decomposition\cite{kessy2018optimal} and Matting Laplacian\cite{levin2007closed} loss to guarantee feature and pixel affinity. However, these 2D style transfers lack spatial consistency and 3D scene perception. On the contrary, our method is capable of maintaining multi-view and cross-time consistency. 

\subsection{3D Stylization}3D style transfer methods aim to synthesize stylized novel views with multi-view consistency. Recently, a large number of 3D style transfer methods based on Neural Radiance Fields(NeRFs)\cite{li2022neural} backbone have emerged, guided by images\cite{huang2022stylizednerf,chiang2022stylizing,nguyen2022snerf,zhang2022arf,liu2023stylerf}, or text\cite{wang2022clip,haque2023instruct,wang2023nerf,park2023ed}. SNeRF\cite{nguyen2022snerf} and ARF\cite{zhang2022arf} achieve visually impressive 3D stylization via NeRF optimization but require time-consuming training for every reference style image. StylizedGS\cite{chiang2022stylizing} introduces NeRF to 3D stylization and trains a hypernetwork for predicting style information of arbitrary style images. Recently, StyleRF\cite{liu2023stylerf} designed a transformation matrix for the 3D feature grid, ensuring capturing style details such as textures and generalization on different styles. However, the inference of NeRF-based stylization is slow, and real-time rendering is unable to be realized. 

Recently, 3D Gaussian Splatting\cite{kerbl20233d} has gained huge popularity due to its high-quality and real-time rendering. And some works have made use of the GS backbone to perform scene stylization. GSS\cite{saroha2024gaussian} and StylizedGS \cite{zhang2024stylizedgs} build up on a 3D-GS backbone for image-based scene stylization, but they require extra optimization for each style image reference. StyleGuassian\cite{liu2024stylegaussian} proposed a 3D style transfer method that can transfer the style of any image to reconstructed 3D-GS instantly. By embedding the Gaussian with VGG\cite{simonyan2014very} features, StyleGaussian can achieve stylization straightly by aligning channel-wise mean and variance with those of the style images in a zero-shot mode. However, StyleGaussian focuses on 3D static scenes, failing to handle the 4D dynamic scene well. Our method is capable of achieving zero-shot high-quality stylization with multi-view and cross-time consistency based on the fast 4D-GS backbone.

\begin{figure*}[!t]
    \centering
    \includegraphics[width=0.85\textwidth]{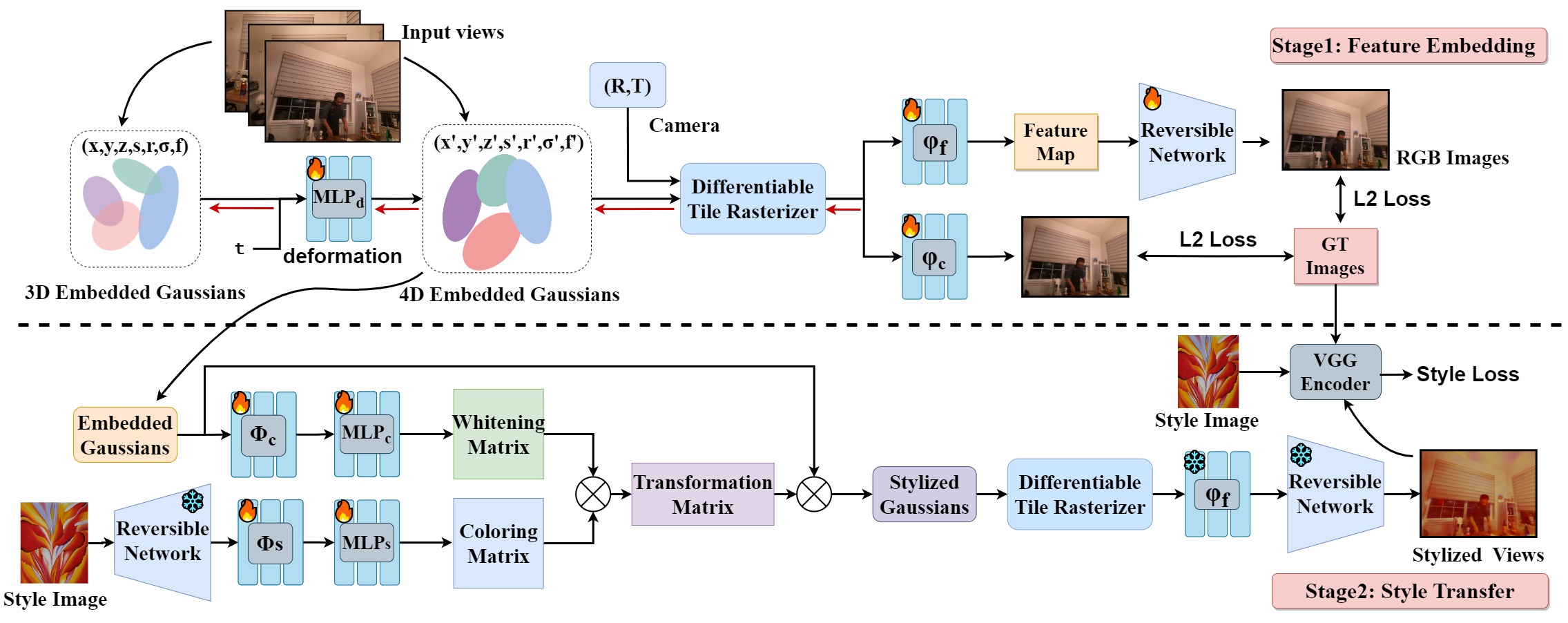}
    \vspace{-0.3cm}
    \caption{\textbf{Overview of our method.} We propose 4DStyleGaussian, a zero-shot style transfer method with 4D Gaussian splatting. Our method pipeline comprises two training stages. Firstly, we train embedded Gaussians with a learnable reversible neural network that ensures preserving the content affinity and clear details. Secondly, we train a linear 4D style transformation matrix with the embedded Gaussians optimized in the first training stage to conduct spatially and temporally consistent style transfer on 4D dynamic scenes.}
    \vspace{-0.4cm}
    \label{framework}
\end{figure*}


\section{Preliminary}
4D Gaussian Splatting (4D-GS)\cite{wu20244d} extends 3D Gaussian Splatting (3D-GS)\cite{kerbl20233d} by constructing a deformation field $\mathcal{D} $ to model both Gaussian motions and shape deformations. Given a camera view $(R,T)$, the rendered pixel color is computed by:
\begin{small}
\vspace{-0.1cm}
\begin{equation}
    c' = \mathcal{G} (\mathcal{S}'|(R,T))
    \label{eq-1}
\vspace{-0.1cm}
\end{equation}
\end{small}
Where $\mathcal{G}$ refers to differential splatting\cite{yifan2019differentiable}. $\mathcal{S}'=[\mathcal{X}',s',r',\sigma, \mathcal{C}]$ and its computation at time $t$ is:
\vspace{-0.2cm}
\begin{small}
\begin{gather}
   \mathcal{S}' = \mathcal{D}(\mathcal{S},t)\; \\ 
    \mathcal{S}=[\mathcal{X},s,r,\sigma, \mathcal{C}]\; 
\vspace{-0.1cm}
\end{gather}
\end{small}
Where $\mathcal{X}=[x,y,z]$ is the center point of each Gaussian. $s$ and $r$ are respectively scaling matrix and rotation matrix that decompose the covariance matrix $\Sigma$ of each Gaussian. $\sigma$ is the opacity and $\mathcal{C}$ is the SH coefficient that models the color of each 3D Gaussian.

Specifically, the color of each pixel is computed by blending N ordered points that overlap the pixel:

\vspace{-0.3cm}
\begin{small}
\begin{equation}
    C = \sum_{i\in N}(c_{i}\alpha_i\prod_{j=1}^{i-1}(1-\alpha_j))
    \label{eq-4}
\end{equation}
\end{small}
Where $c_{i}$ represents the color of each Gaussian that is calculated with the SH coefficient  $\mathcal{C}$. $\alpha_{i}$ is given by evaluating a 2D Gaussian with covariance $\Sigma$ multiplied with a learned per-point opacity $\sigma_i$.

In order to assign the Gaussians with deep features that are typically used for the style transfer task, we aim to substitute the color $c_i$ with the feature $f_i$, which is to be discussed in the next section.

\section{4DStyleGaussian} In this section, we propose 4DStyleGaussian, a zero-shot style transfer method for 4D Gaussian Splatting. The pipeline of our method is illustrated in Fig.\ref{framework}, comprising two training stages. Firstly, with a learnable reversible neural network, we train an embedded Gaussian Splatting where each Gaussian is endowed with a high-dimension feature. Then, we train a linear 4D style transformation matrix for spatially and temporally consistent style transfer with the embedded Gaussians optimized in the first training stage and the style images. 

\subsection{4D Embedded Gaussians}\label{subsection:4D Embedded Gaussians}
\subsubsection{High-dimension Feature Rendering.}
We use the rendering pipeline based on the 4D differentiable Gaussian Splatting framework designed by 4D-GS\cite{wu20244d}. We follow the multi-resolution neural voxels to establish relationships between 3D Gaussians, which encode deformations within the unit voxel grid based on the neighboring points. Then a small MLP is used to decode the voxel feature and get $\mathcal{S'}$. Practically, we change the original $\mathcal{S'}=[\mathcal{X}',s',r',\sigma,\mathcal{C}]$ to $\mathcal{S'}=[\mathcal{X}',s',r',\sigma,\mathcal{F}]$, where $\mathcal{C} \in \mathbb{R}^3$ is the original SH coefficient that models the color of each Gaussian in 4D-GS\cite{wu20244d}. $\mathcal{F} \in \mathbb{R}^{32}$ is the SH coefficient of the feature $f_i$ that embeds each Gaussian. 

By projecting the 3D Gaussain to get 2D covariance $\Sigma'$, the embedded feature map $E \in  \mathbb{R}^{32}$ are computed by differentiable volumetric rendering:

\begin{small}
\vspace{-0.4cm}
\begin{gather}
    \alpha_{i} = \sigma_ie^{-\frac{1}{2}(p-\mu_i)^T\Sigma'(p-\mu_i)}\;\\
    \vspace{-0.3cm}
    E = \sum_{i\in N}(f_{i}\alpha_{i}\prod_{j=1}^{i-1}(1-\alpha_{j}))\;
\vspace{-0.4cm}
\end{gather}
\end{small}
Where $\sigma_i$ is the opacity of each Gaussian. $\mu_i$ is the uv coordinates of the 3D Gaussians projected onto the 2D image plane. $p$ is the pixel in the image plane. More details on the calculation are provided in supplementary materials.

To guarantee independent optimization for both the color map $C$ and feature map $F$, we utilize two different MLPs($\psi_c$ and $\psi_f$) to transform the rendered embedding feature maps to color map and feature map $F$:
\begin{small}
\vspace{-0.2cm}
\begin{gather}
   C = \psi_c(E)\; \label{eq6}\\ 
   F = \psi_f(E)\; 
   \label{eq7}
\vspace{-0.3cm}
\end{gather}
\end{small}

\subsubsection{Lossless Feature Embedding.}
Existing 3D style transfer methods typically first distill the feature maps from multi-view images with a pre-trained encoder such as VGG into the 3D representation such as NeRF or Gaussian Splatting. Then in the feature space, they conduct style transfer and get the stylized feature maps of multi-views which are then decoded back to RGB images. However, the traditional encoders are designed for object-level classification and this character makes it prominently cause artifacts and much content information dropout. To make the Gaussians capable of preserving pixel affinity, we use a reversible neural network for lossless feature embedding to train the embedded Gaussians. The reversible neural network consists of specifically designed reversible residual networks that can not only preserve content affinity but reduce redundant information by taking advantage of the bijective transformation. To train the embedded Gaussians, we minimize the loss as follows:
\begin{small}
\begin{equation}
    L_c = \lambda_c\| C  - I_{gt}\|_2^2 +  \lambda_f\| R_{rev}(F)  - I_{gt}\|_2^2
\end{equation}
\end{small}
$R_{rev}$ refers to the reverse process of the reversible neural network and $I_{gt}$ is the ground truth images.

\begin{figure*}[!t]
    \centering
    \includegraphics[width=0.8\textwidth]{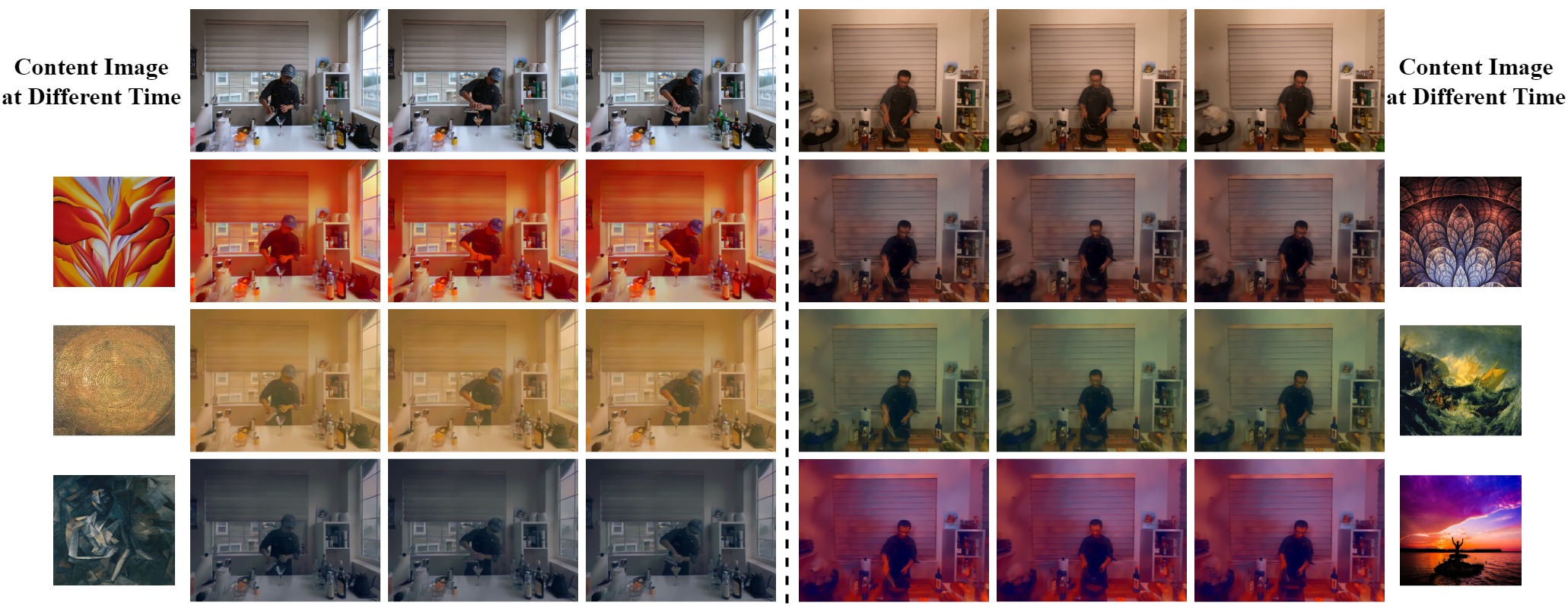}
    \vspace{-0.3cm}
    \caption{\textbf{Qualitative results of stylized novel views at different times and styles.} Our method can generate high-quality stylized synthesis and have good generalization on various style images while maintaining spatial-temporal consistency.}
    \label{ours-qualitative}
    \vspace{-0.4cm}
\end{figure*}

\subsection{4D Style Transfer}\label{subsection:4D Style Transfer}
In this section, we aim to find a linear style transformation matrix that is consistent in 4D feature space to transform the embedded Gaussians according to arbitrary style images, ensuring stylization with multi-view and cross-time consistency.

Denote that $G_E\in \mathbb{R}^{N_G\times32}$ is the embedded features of all Gaussians. $N_G$ is the number of Gaussian points. $\mathcal{R}_{f}$ refers to the forward process of the pre-trained reversible neural network. $I_s$ is the style image. We utilize an MLP $\Phi_c$ to extract the content feature $F_c$ from the embedded Gaussians $G_E$ and a 2D CNN network $\Phi_s$ to obtain the style feature $F_s$:
\vspace{-0.1cm}
\begin{small}
\begin{gather}
    F_c=\Phi_{c}(G_E);\ \\
    F_s=\Phi_{s}(\mathcal{R}_{f}(I_s));\
\vspace{-0.3cm}
\end{gather}
\end{small}
\vspace{-0.4cm}

The goal is to find a transformation matrix $T=T_sT_c$ that can transform the content feature $F_c$ to $F_{cs}$ which is consistent with the style feature space $F_s$, and it is optimized by minimizing the loss:
\begin{figure*}[!t]
    \centering
    \includegraphics[width=0.8\textwidth]{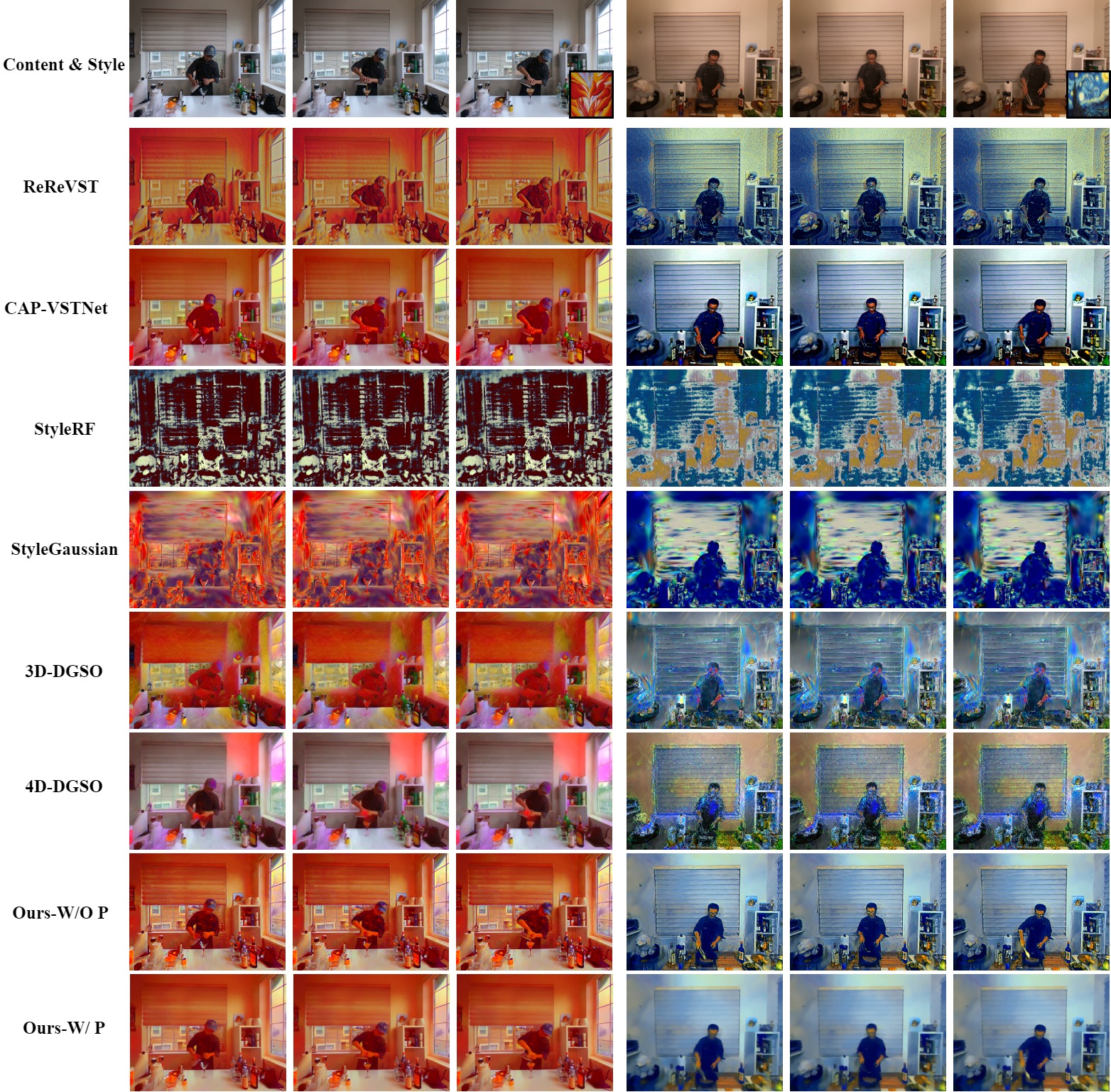}
    \vspace{-0.3cm}
    \caption{\textbf{Comparsions with baselines of stylized novel views at different times and styles.} Our method has better performance than the baselines, achieving better visual effects, and is consistent with the style input with spatial-temporal consistency.}
    \vspace{-0.5cm}
    \label{baseline-qualitative}
\end{figure*}
\begin{small}
\vspace{-0.2cm}
\begin{equation}
    L = \frac{1}{N_f}\| \overline{F}_{cs} \overline{F}_{cs}^T  -\overline{F}_s \overline{F}_s^T  \|_2^2
\vspace{-0.2cm}
\end{equation}
\end{small}
Where $\overline{F}_{cs}= T(F_c-\mu_c)$ and $\overline{F}_{s}= F_s-\mu_s$. $\mu_c$ is the mean value of $F_c$. $\mu_s$ is the mean value of $F_s$. $N_f$ is the dimension in the feature space. 

Through Singular Value Decomposition(SVD), the closed-form solution of $T^*$ can be calculated by:
\vspace{-0.2cm}
\begin{small}
\begin{gather}
    T^{*}\overline{F}_{c}\overline{F}_{c}^TT^{*T} = \overline{F}_{s}\overline{F}_{s}\;\\ \label{T1}
    \overline{F}_{c}\overline{F}_{c}^T=W_c\Sigma_cW_c^T\;\\ \label{F_c}
      \overline{F}_{s}\overline{F}_{s}=W_s\Sigma_sW_s^T\;\\
    T^*=T_sT_c=\left(W_s\Sigma_s^{\frac{1}{2}}W_s^T\right)\left(W_c\Sigma_c^{\frac{1}{2}}W_c^T\right) \; \label{T2}
\vspace{-0.6cm}
\end{gather}
\end{small}

From the equations above, the whitening matrix $T_c$ is only related to the content feature, and the coloring matrix $F_s$ is only related to the style feature $F_s$. Consequently, we design two different MLPs to optimize $T_c$ and $T_s$ from $F_c$ and $F_s$ respectively: 

\begin{small}
\vspace{-0.3cm}
\begin{equation}
    T_c=\text{MLP}_c(\pi(F_c))
    \label{eq14}
\vspace{-0.2cm}
\end{equation}
\end{small}
\vspace{-0.25cm}
\begin{small}
\begin{equation}
    T_s=\text{MLP}_{s}(\pi(F_s))
    \label{eq15}
\vspace{-0.15cm}
\end{equation}
\end{small}
Where $\pi$ is a flattening operation for reshaping $F_c$ and $F_s$ to the matrix shape.

The computation of the stylized feature map $\hat{F}_{cs}$ as follows:

\begin{small}
\vspace{-0.5cm}
\begin{equation}
    f_i^{cs}=T_sT_c(f_i-\mu_f)+\mu_s \label{eq16}
\end{equation}
\end{small}
\vspace{-0.4cm}
\begin{small}
\begin{equation}
    F_{cs}=\psi_f\left(\sum_{i=1}^{N} \left( f_i^{cs} \alpha_{i}\prod_{j=1}^{i-1}(1-\alpha_{j})\right)\right)\label{eq17}
\vspace{-0.1cm}
\end{equation}
\end{small}
Where $\mu_f=\mu(G_E)$. $\mu_s =\mu(\mathcal{R}_f(I_s))$. $f_i$ is the embedded feature of each Gaussian. $\psi_f$ is the pre-trained MLP in Eq.\ref{eq7}.
With rendered feature map $F$ calculated with Eq.\ref{eq7}, Eq.\ref{eq16} and Eq.\ref{eq17} can be rewritten to:

\begin{small}
\vspace{-0.6cm}
\begin{equation}
    \begin{split}
    F_{cs}&=T_sT_c\left(\psi_f\left(\sum_{i=1}^{N}f_i\alpha_{i}\prod_{j=1}^{i-1}(1-\alpha_{j})\right)-\mu_f\right)+\mu_s \\ 
    &=T_sT_c(F-\mu_f)+\mu_s
    \end{split}
    \label{eq18} 
\vspace{-0.2cm}
\end{equation}
\end{small}

With Eq.\ref{eq14},\ref{eq15},\ref{eq18}, we can obtain the stylized feature ${F}_{cs}$, with which we reverse the high-dimension feature back to RGB space and get the stylized novel views:
\begin{small}
\vspace{-0.2cm}
\begin{equation}
    \hat I_{trans} =R_{rev}(F_{cs})
    \label{eq20}
\vspace{-0.1cm}
\end{equation}
\end{small}

To make the stylizations preserve content information from original content images while presenting style patterns correlated with visual semantics of style instructions, we design the following art loss:
\begin{small}
\vspace{-0.2cm}
\begin{equation}
    L_{content} =\| \Phi_{enc}(\hat I_{trans})  - \Phi_{enc}(I_{gt})  \|_2^2 
    \label{eq20}
\end{equation}
\end{small}
\vspace{-0.5cm}
\begin{small}
\begin{equation}
\begin{aligned}
    L_{style} =\| \mu(\Phi_{enc}(\hat I_{trans}) ) - \mu(\Phi_{enc}(I_{s}))  \|_2^2 +\\
    \| \sigma(\Phi_{enc}(\hat I_{trans}))  - \sigma(\Phi_{enc}(I_{s}) ) \|_2^2
    \label{eq21}
\end{aligned}
\end{equation}
\end{small}

\vspace{-0.1cm}
\begin{small}
\vspace{-0.1cm}
\begin{equation}
    L_{s} =  \lambda_{c} * L_{content} + \lambda_{s} * L_{style}  
    \label{eq23}
\end{equation}
\vspace{-0.1cm}
\end{small}
Where $\Phi_{enc}$ is a pre-trained VGG network. 

\begin{table*}[h]

\centering
\tiny
\begin{tabular}{c|cc|cc|cc|cc|cc|cc|cc}
\hline
\multirow{2}{*}{\textbf{Method}} & \multicolumn{2}{c|}{\textbf{\emph{Cook\_spinach}}} & \multicolumn{2}{c|}{\textbf{\emph{Flame\_steak}}} & \multicolumn{2}{c|}{\textbf{\emph{Sear\_steak}}} & \multicolumn{2}{c|}{\textbf{\emph{Flame\_salmon\_1}}} & \multicolumn{2}{c|}{\textbf{\emph{Coffee\_martini}}} & \multicolumn{2}{c|}{\textbf{\emph{Cut\_roasted\_beef}}}  & \multicolumn{2}{c}{\emph{\textbf{Mean}}} \\ 
            & \textbf{RMSE}         & \textbf{LPIPS}      & \textbf{RMSE}     &\textbf{LPIPS}  & \textbf{RMSE}      & \textbf{LPIPS}      & \textbf{RMSE}  & \textbf{LPIPS}       & \textbf{RMSE}      & \textbf{LPIPS}   &\textbf{RMSE}     & \textbf{LPIPS}    & \textbf{RMSE}     & \textbf{LPIPS}        \\ \hline \hline
StyleRF  &0.050 &0.025 &0.048 &0.019 &0.045 &0.016 &0.026 &0.011 &0.037 &0.014 &0.022 &0.023 &0.038 &0.018            \\ \hline

ReReVST  &0.080 &0.101 &0.077 &0.090 &0.076 &0.090 &0.059 &0.069 &0.060 &0.073 &0.080 &0.099 &0.072 &0.087            \\ \hline
CAP-VSTNet &0.053 &0.060 &0.050 &0.058 &0.053 &0.058 &0.042 &0.042 &0.044 &0.045 &0.052 &0.060 &0.049 &0.054            \\ \hline

3D-DGSO &0.040 &\textbf{0.017} &0.040 &0.020 &0.040 &0.018 &0.035 & 0.011 &0.041 &0.013 &0.047 &0.030 &0.041 &0.018            \\ \hline

StyleGaussian   &0.035 &{0.018} &0.033 &\textbf{0.012} &0.035 &0.013 &0.025 &0.008 &0.021 &0.008 &0.035 &0.019 &0.031 &0.013 \\         \hline 
4D-DGSO &0.054 &0.029 &0.049 &0.024 &0.049 &0.021 &0.042 &0.018 &0.021 &0.007 &0.060 &0.034 &0.046 &0.022      \\         \hline \hline

Ours  &\textbf{0.015} &\textbf{0.017} &\textbf{0.020} &{0.015} &\textbf{0.013} &\textbf{0.009} &\textbf{0.009} &\textbf{0.005} &\textbf{0.009} &\textbf{0.005} &\textbf{0.019} &\textbf{0.018} &\textbf{0.015} &\textbf{0.012}               \\     \hline          
\end{tabular}
\vspace{-0.3cm}
\caption{\textbf{Quantitative comparisons on short-range consistency.} We compare the consistency scores of LPIPS ($\downarrow$) and RMSE ($\downarrow$) between stylized images at 2 adjacent novel views to evaluate the short-range consistency. The best performance is in \textbf{bold}.}
\vspace{-0.2cm}
\label{tab:short-range-consistency}
\end{table*}

\begin{table*}[h]
\centering
\tiny
\begin{tabular}{c|cc|cc|cc|cc|cc|cc|cc}
\hline
\multirow{2}{*}{\textbf{Method}} & \multicolumn{2}{c|}{\textbf{\emph{Cook\_spinach}}} & \multicolumn{2}{c|}{\textbf{\emph{Flame\_steak}}} & \multicolumn{2}{c|}{\textbf{\emph{Sear\_steak}}} & \multicolumn{2}{c|}{\textbf{\emph{Flame\_salmon\_1}}} & \multicolumn{2}{c|}{\textbf{\emph{Coffee\_martini}}} & \multicolumn{2}{c|}{\textbf{\emph{Cut\_roasted\_beef}}}  & \multicolumn{2}{c}{\emph{\textbf{Mean}}} \\ 
            & \textbf{RMSE}         & \textbf{LPIPS}      & \textbf{RMSE}     &\textbf{LPIPS}  & \textbf{RMSE}      & \textbf{LPIPS}      & \textbf{RMSE}  & \textbf{LPIPS}       & \textbf{RMSE}      & \textbf{LPIPS}   &\textbf{RMSE}     & \textbf{LPIPS}    & \textbf{RMSE}     & \textbf{LPIPS}        \\ \hline \hline

StyleRF &0.058 &0.035 &0.057 &0.030 &0.051 &0.019 &0.029 &0.013 &0.047 &0.022 &0.055 &0.033 &0.050 &0.025            \\ \hline

ReReVST &0.083 &0.107 &0.078 &0.094 &0.077 &0.091 &0.060 &0.072 &0.063 &0.077 &0.081 &0.103 &0.074 &0.091            \\ \hline
CAP-VSTNet  &0.058 &0.070 &0.055 &0.065 &0.064 &0.071 &0.044 &0.044 &0.050 &0.050 &0.055 &0.068 &0.087 &0.105            \\ \hline

3D-DGSO    &0.045 &\textbf{0.020} &0.049 &0.028 &0.045 &0.021 &0.041 &0.014 &0.052 &0.022 &0.052 &0.035 &0.047 &0.023          \\ \hline

StyleGaussian  &0.049 &{0.030} &0.049 &\textbf{0.022} &0.040 &\textbf{0.015} &0.030 &{0.010} &0.032 &0.015 &0.037 &{0.025} &0.039 &{0.019 }     \\         \hline 
4D-DGSO   &0.064 &0.046 &0.057 &0.034 &0.057 &0.029 &0.054 &0.027 &0.029 &0.012 &0.066 &0.046 &0.054 &0.033\\         \hline \hline

Ours    &\textbf{0.022} &0.025 &\textbf{0.025} &{0.024} &\textbf{0.020} &{0.017} &\textbf{0.009} &\textbf{0.006} &\textbf{0.013} &\textbf{0.008} &\textbf{0.023} &\textbf{0.023} &\textbf{0.020} &\textbf{0.018}                \\     \hline          
\end{tabular}
\vspace{-0.3cm}
\caption{\textbf{Quantitative comparisons on long-range consistency.} We compare the consistency scores of LPIPS ($\downarrow$) and RMSE ($\downarrow$) between stylized images at 4 adjacent novel views to evaluate the long-range consistency. The best performance is in \textbf{bold}.}
\vspace{-0.4cm}
\label{tab:long-range-consistency}
\end{table*}

\subsection{Photo-realistic Propagation}
\vspace{-0.1cm}
Though applying a reversible neural network to train the embedded Gaussians can help reduce the artifacts and content loss that are relatively severe in previous pre-trained deep encoder-decoder frameworks, the original stylization results still have some artifacts. It may happen when the Gaussian Splatting creates elongated artifacts or “splotchy” Gaussians due to anisotropic Gaussians, or visibility algorithm which leads to Gaussians suddenly changing depth/blending orders. 

To solve the problem, We apply a convolution spatial propagation network $\mathcal{P}$ for decreasing distortions. The network takes reconstructed images from embedded Gaussian $R_{rev}(F)$ as conditional input and filters the original stylization results by Eq.\ref{eq24}:
\begin{small}
\begin{equation}
    \hat I_{pro} =\mathcal{P}( \hat I_{trans},\text{Whiten}(R_{rev}(F)))
    \label{eq24}
\end{equation}
\end{small}
Where $\text{Whiten}(\cdot)$ is a 2D whitening operation. We train the propagation module by:
\vspace{-0.2cm}
\begin{small}
\begin{equation}
    L_{pro} =\| \Phi_{enc}(\hat I_{pro})  - \Phi_{enc}(R_{rev}(F))  \|_2^2 
    \label{eq25}
\end{equation}
\end{small}

\section{Experiment}
\subsection{Implementation Details}

\subsubsection{Dataset and Experiment settings.} We utilize the dataset provided by DyNeRF\cite{li2022neural} to evaluate our method. DyNeRF's dataset is captured using 21 static cameras and at a frame rate of 30 FPS. Practically, We use 20 views for training and 1 view for testing. We utilize WikiArt\cite{wikiart2018visual} as the style images dataset. Our implementation is based on the PyTorch framework and tested on one RTX 3090Ti GPU. The feature embedding stage takes 3000 iterations for the coarse stage which optimizes static 3D Gaussians, and 1.5k iterations for the fine stage of optimizing 4D Gaussians while the style transfer training stage takes 1.5k iterations. The whole training process takes about 6 hours and the inferencing time is about 0.037s for one frame.
\begin{figure}[!t]
\centering
    \includegraphics[width=0.9\columnwidth]{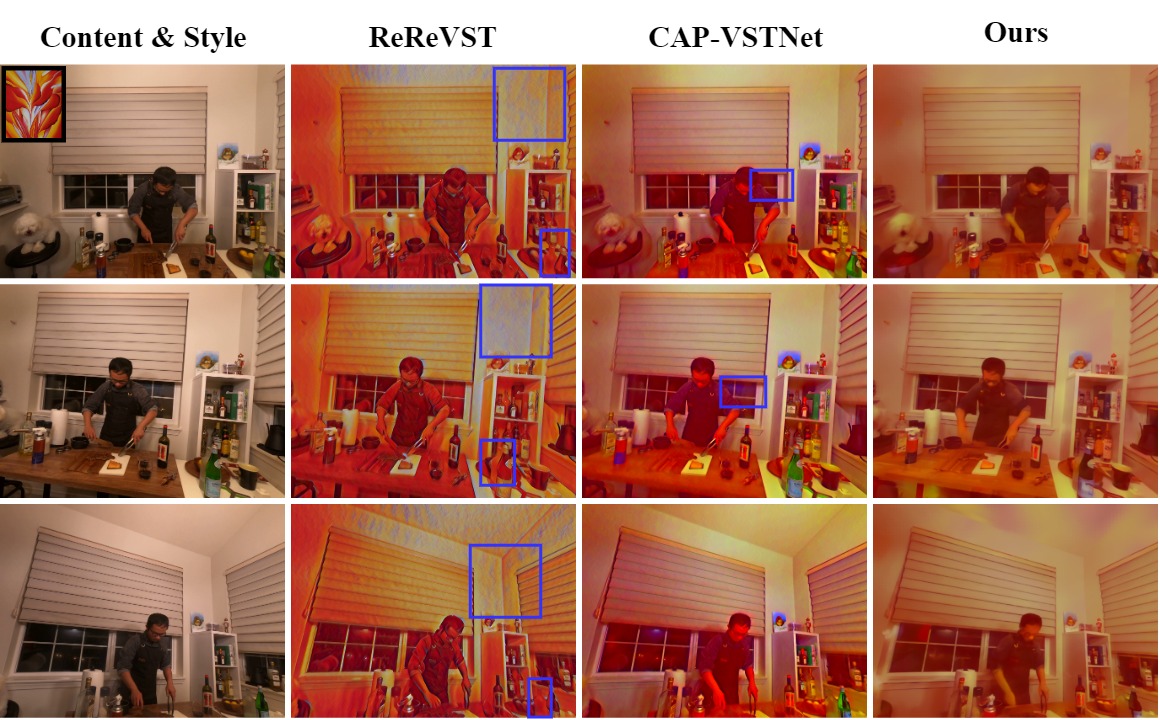}
    \vspace{-0.2cm}
    \caption{\textbf{Comparsions of stylized novel views at different times with video transfer methods.} CAP-VSTNet and ReReVST show inconsistency between different views of different video frames(As shown in the blue box in the figures) while our method illustrates better consistency between multiple views and across time.}
    \vspace{-0.5cm}
    \label{video_com}
\end{figure}


\subsubsection{Baselines.} We compare our method with StyleRF\cite{liu2023stylerf}, StyleGaussian\cite{liu2024stylegaussian} and DGSO\cite{kerbl20233d}. StyleRF is a zero-shot 3D style transfer method with NeRFs and StyleGaussian optimizes a style transformation matrix with GS. DGSO directly optimized GS with a specific style input. Since there is no previous work of style transfer for 4D dynamic scenes based on Gaussian Splatting, we directly optimize a 4D Gaussian Splatting with the input style images following DGSO\cite{kerbl20233d}. Additionally, we also compare our method with video style transfer methods: CAP-VSTNet\cite{wen2023cap} and ReReVST\cite{wang2020consistent}.

\subsection{Qualitative Results}
\subsubsection{Zero-shot Stylization}
Fig.\ref{ours-qualitative} shows that our method can generate high-quality stylized synthesis and have good generalization on various styles while maintaining multi-view and cross-time consistency.

\subsubsection{Comparision With Baselines}
Fig.\ref{baseline-qualitative} illustrates the qualitative comparisons with the baselines on the Dynerf dataset. ReReVST produces structural distortions and contains apparent artifacts. CAP-VSTNet can not realize reasonable stylization consistent with the style images. StyleRF, StyleGaussian and 3D-DGSO fail to reconstruct the dynamic scenarios and stylizations contain many artifacts and distortions. 4D-DGSO has many artifacts caused by the limitation of Gaussian Splatting, and simply optimizing a stylized 4D-GS might worsen the distortions. When omitting a photo-realistic propagation module(As shown in "Ours-W/O P" in Fig.\ref{baseline-qualitative}), our method appears to have some artifacts caused by the Gaussians. With the photo-realistic propagation optimization, our method has better visual effects and is consistent with the style input with spatial-temporal consistency.

Fig.\ref{video_com} demonstrates a further qualitative comparison with video style transfer methods. We choose three camera views and three time stamps. CAP-VSTNet and ReReVST can handle consistency between views across a video frame, but they show inconsistency between different views of different video frames(As shown in the blue box in the figure). This is because video style transfer methods lack spatial consistency and 3D scene perception.

\begin{table}[h]
\centering

\label{tab:time}
\small
\begin{tabular}{c|c|c}
\hline
Method    & Inferencing Time(s)   & Styles\\ 
\hline \hline
StyleRF &  $30.22$ &$ \infty$\\ 
\hline 
ReReVST & $0.155$  & $\infty$ \\ 
\hline 
CAP-VSTNet  & $0.231 $ & $\infty$\\ 
\hline
StyleGaussian &$0.189$  &$ \infty$\\
\hline
3D-DGSO & $0.003$  & $1$\\
\hline
4D-DGSO & $0.027$  &$ 1$ \\
\hline \hline
Ours   & $0.037$ &$ \infty$\\ 
\hline
\end{tabular}
\vspace{-0.2cm}
\caption{\textbf{Comparsion of Inferencing Efficiency.} Our method significantly has better performance than the other methods in terms of time efficiency.}
\vspace{-0.2cm}
\label{tab:time}
\end{table}

\vspace{-0.3cm}
\begin{figure}[h]
\centering
\includegraphics[width=0.8\columnwidth]{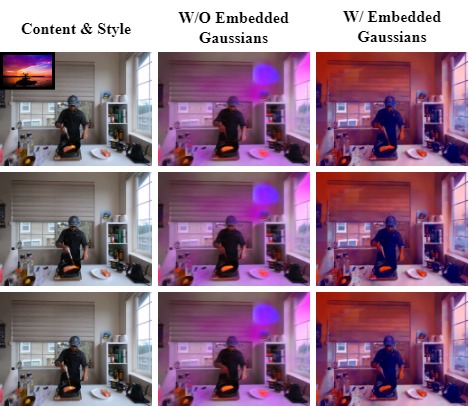}
\vspace{-0.2cm}
\caption{\textbf{Ablation study of embedded Gaussians.} The results present that training embedded Gaussians achieves better stylization without many artifacts and distortions.}
\vspace{-0.3cm}
\label{4D-GS}
\end{figure}

\begin{figure}[h]
\vspace{-0.2cm}
\centering
\includegraphics[width=0.8\columnwidth]{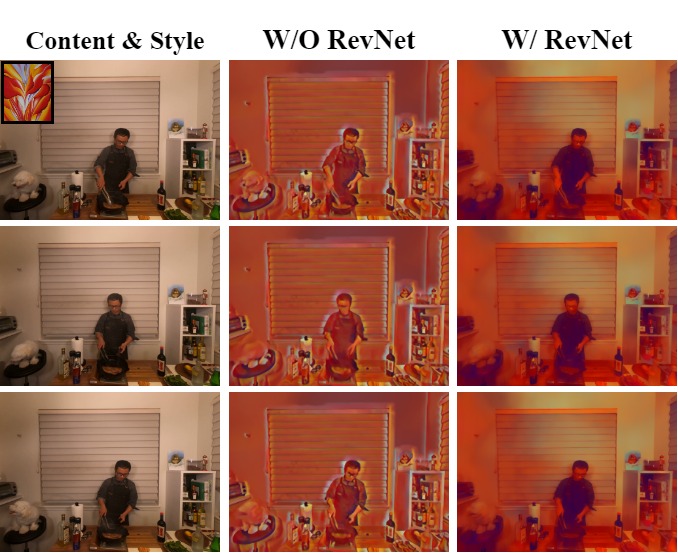}
\vspace{-0.2cm}
\caption{\textbf{Ablation study of Reversible Neural Network.} Applying a reversible neural network is capable of reducing unnecessary artifacts and increasing the details of the stylized images. This is because there is a down-sample when using the VGG encoder to get the feature map and this will inevitably lead to blur or artifacts.}
\vspace{-0.5cm}
\label{rev}
\end{figure}
\vspace{-0.3cm}
\begin{figure}[h]
\centering
\includegraphics[width=0.8\columnwidth]{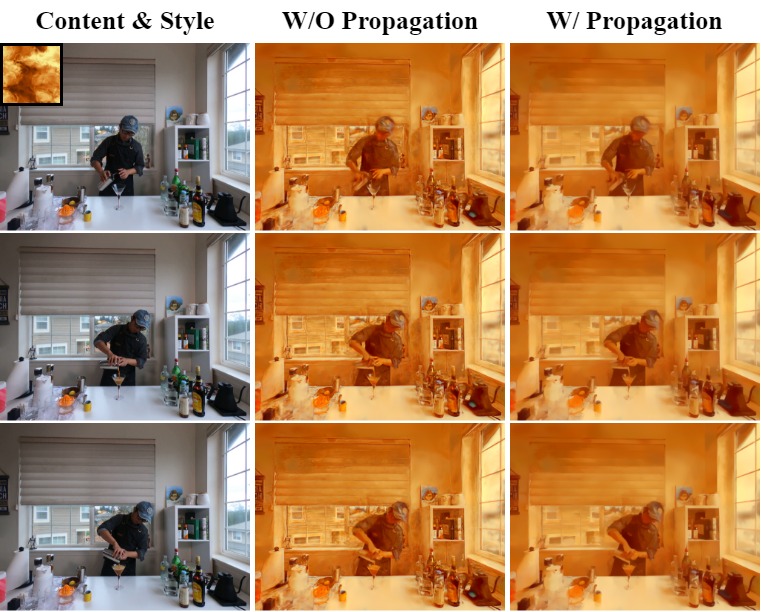}
\vspace{-0.2cm}
\caption{\textbf{Ablation study of photo-realistic propagation.} Adopting a photo-realistic propagation module helps obtain better stylization results without artifacts and distortions.}
\vspace{-0.4cm}
\label{propagation}
\end{figure}

\subsection{Quantitative Results}
\subsubsection{Efficiency}
As shown in Tab.\ref{tab:time}, "Time of Inference" refers to transferring time and rendering time. "$\infty$" means it can transfer every style without extra optimization. 3D-DGSO and 4D-DGSO need less inferencing time, but they require training for each style input. Our method significantly has better performance than the other methods.  
\vspace{-0.2cm}
\subsubsection{Consistency}
We compute the short-range and long-range consistency as the quantitative metrics. We first transform the stylized novel views from one view to another view by utilizing a pre-computed optical flow. Then we evaluate the LPIPS and RMSE of the two views to assess the consistency. The results are shown in Tab.\ref{tab:short-range-consistency} and Tab.\ref{tab:long-range-consistency}, showing that our method achieves better stylizations with multi-view and cross-time consistency than the baselines.

\subsection{Ablation Study}

\subsubsection{4D Embedded Gaussians}

As shown in Fig.\ref{4D-GS}, "W/O Embedded Gaussians" refers to directly optimizing a stylized 4D Gaussian Splatting based on a style image. The results present that training embedded Gaussians achieves better stylization without many artifacts and distortions which can be found in the result of "W/O Embedded Gaussians".
\vspace{-0.2cm}

\subsubsection{Reversible Neural Network}
For the ablation of whether to use the reversible neural network, we conduct another experiment that distills the VGG feature into the Gaussians. As shown in Fig.\ref{rev}, applying a reversible neural network is capable of reducing unnecessary artifacts and increasing the details of the stylized images. This is because there is a down-sample when using the VGG encoder to get the feature map and this will inevitably lead to blur or artifacts.

\vspace{-0.2cm}
\subsubsection{Photo-realistic Propagation}
As shown in Fig.\ref{propagation}, adopting a photo-realistic propagation module helps obtain better stylization results without artifacts and distortions, which are caused by the elongated artifacts or “splotchy” Gaussians due to anisotropic Gaussians. 

\vspace{-0.2cm}
\subsection{Style Interpolation}
\begin{figure}[h]
    \centering
    \includegraphics[width=0.9\columnwidth]{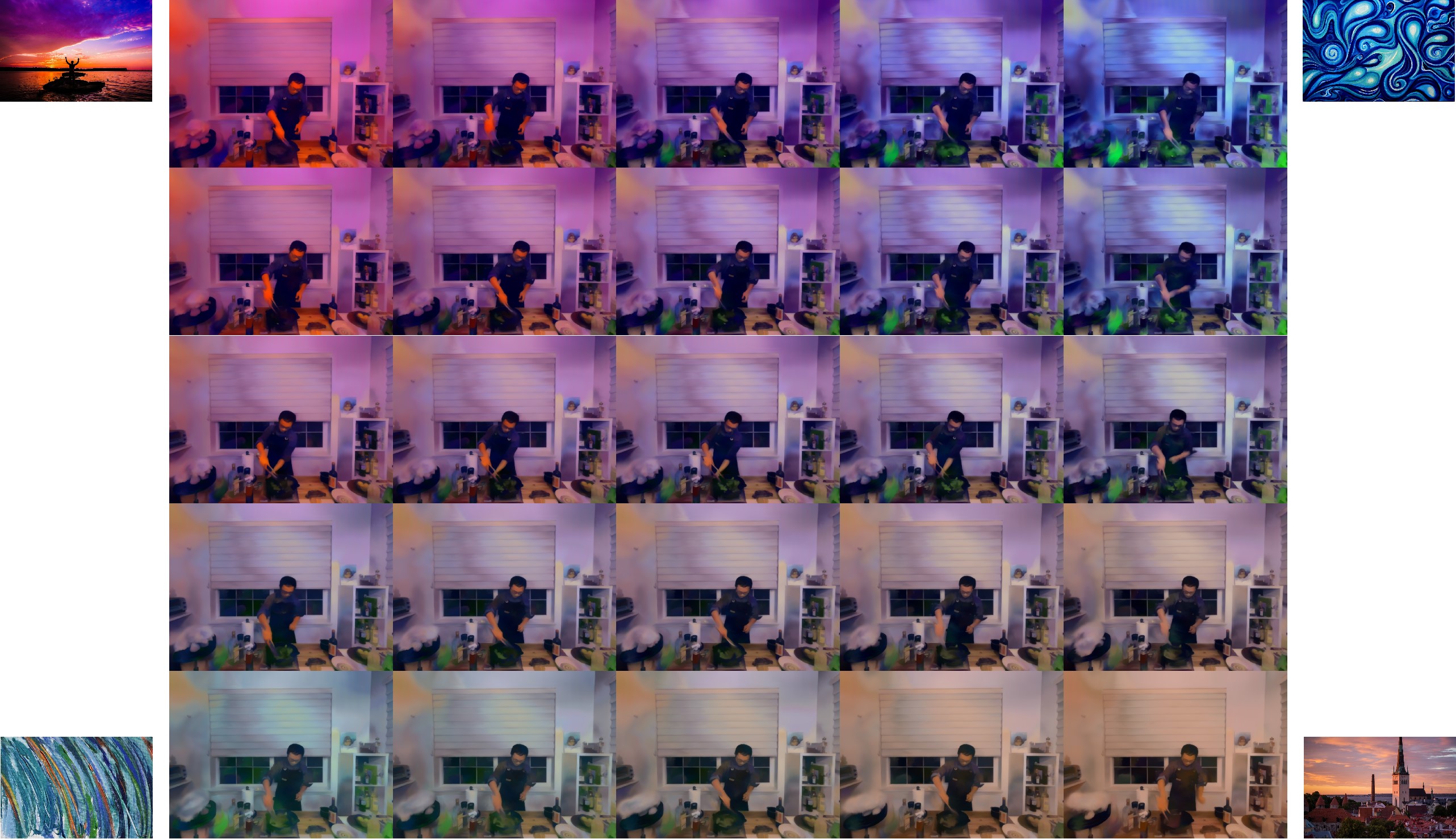}
    \vspace{-0.3cm}
    \caption{\textbf{Results of style interpolation.} Our method can have a smooth multi-style interpolation between different styles and different times since the 4D style transformation matrix is linear.}
    \vspace{-0.5cm}
    \label{interpolation}
\end{figure}

Fig.\ref{interpolation} demonstrates the 4D stylization results of interpolation among four different style images. The rendered stylized novel views are in chronological order. Our method can have a smooth multi-style interpolation between different styles and different times since the 4D style transformation matrix is linear. 
\vspace{-0.2cm}
\vspace{-0.3cm}
\section{Conclusion}

In this paper, we introduce 4DStyleGaussian, a zero-shot 4D style transfer method with Gaussian Splatting that achieves real-time stylization of arbitrary style images. We train embedded Gaussians in 4D space with a reversible neural network to ensure content affinity, which significantly reduces the artifacts caused by traditional encoders used for feature distillations. With the trained embedded Gaussians, we optimize a 4D style transformation matrix that guarantees zero-shot stylizations with multi-view and cross-time consistency. Qualitative and quantitative experiments demonstrate our method outperforms the existing 3D stylization methods in terms of multi-view consistency, cross-time consistency, efficiency and effectiveness.

\bibliography{reference}

\end{document}